\documentclass[a4paper,fleqn]{cas-sc} 
\usepackage{amsmath}
\usepackage{algorithm}
\usepackage{algpseudocode}
\usepackage[utf8]{inputenc}
\usepackage{booktabs} 
\usepackage[numbers]{natbib}

\def\tsc#1{\csdef{#1}{\textsc{\lowercase{#1}}\xspace}}
\tsc{WGM}
\tsc{QE}
\tsc{EP}
\tsc{PMS}
\tsc{BEC}
\tsc{DE}

\begin{document}
\let\WriteBookmarks\relax 
\def\floatpagepagefraction{1} 
\def\textpagefraction{.001} 
\shorttitle{ Long working distance smartphone microscope}
\shortauthors{Zhengang Lu et~al.} 
%\begin{frontmatter}

\title [mode = title]{Long working distance portable smartphone microscopy for metallic mesh defect detection}                      

\author[1,2]{Zhengang Lu}[style=chinese,orcid=0000-0001-6490-5819]

\cormark[1]
\ead{luzhengang@hit.edu.cn}

\author[1,2]{Hongsheng Qin}[style=chinese]
\ead{21B901017@stu.hit.edu.cn}

\author[1,2]{Jing Li}[style=chinese]
\ead{qinhs02@163.com}

\author[3]{Ming Sun}[style=chinese]
\ead{}

\author[1,2]{Jiubin Tan}[style=chinese]

\address[1]{Ultra-precision Optical \& Electronic Instrument Engineering Center, Harbin Institute of Technology, Harbin 150001, China}
\address[2]{Key Lab of Ultra-precision Intelligent Instrumentation (Harbin Institute of Technology), Ministry of Industry and Information Technology, Harbin 150001, China}
%\address[3]{Institute of Materials, China Academy of Engineering Physics, Jiangyou, 621908, Sichuan, China}
\address[3]{Visual Computing Center, King Abdullah University of Science and Technology (KAUST),  Thuwal 23955, Saudi Arabia}

\begin{abstract} 
Metallic mesh is a transparent electromagnetic shielding film with a fine metal line structure. However, it can develop defects that affect the optoelectronic performance whether in the production preparation or in actual use.
The development of in-situ non-destructive testing (NDT) devices for metallic mesh requires long working distances, reflective optical path design, and miniaturization.
To address the limitations of existing smartphone microscopes, which feature short working distances and inadequate transmission imaging for industrial in-situ inspection, we propose a novel long-working distance reflective smartphone microscopy system (LD-RSM). 
LD-RSM builds a 4\textit{f} optical imaging system with external optical components and a smartphone, utilizing a beam splitter to achieve reflective imaging with the illumination system and imaging system on the same side of the sample. It achieves an optical resolution of 4.92$\mu$m and a working distance of up to 22.23 mm.
Additionally, we introduce a dual prior weighted Robust Principal Component Analysis (DW-RPCA) for defect detection. This approach leverages spectral filter fusion and Hough transform to model different defect types, enhancing the accuracy and efficiency of defect identification. Coupled with an optimized threshold segmentation algorithm, DW-RPCA method achieves a pixel-level accuracy of 84.8\%.
Our work showcases strong potential for growth in the field of in-situ on-line inspection of industrial products.
\end{abstract}

%graphicalabstract
%\includegraphics{figs/grabs.pdf}

%highlights
\begin{highlights}
\item  Proposed LD-RSM has a long working distance for non-destructive testing.
\item The light source and imaging system are located on the same side of the sample for industrial in-situ detection.
\item A defect detection model (DW-RPCA) based on spectral filtering fusion and Hough transform is proposed.
\item An imaging system with a detection model for metallic mesh, has been proposed for the first time.
\item The proposed method could emerge as a promising alternative for industrial in-situ testing.
\end{highlights}

%keywords

\maketitle

\section{Introduction}
In recent, rapid development of optoelectronic devices, including communication and medical devices, and detectors, has enhanced human life significantly \cite{li2023silver}. However, these devices emit large amounts of electromagnetic radiation, leading to electromagnetic interference (EMI). 
A well-designed metallic mesh can provide effective EMI shielding while maintaining transparency. \cite{liang2023metal}.
Metallic mesh is formed by periodic structures of thin metal films, achieving high-frequency visible light transmission and low-frequency microwave cutoff by carefully selecting structural unit periods and line widths \cite{lu2024valid}.

At present, there are many methods for the preparation of metallic mesh, such as photolithography coating \cite{qiu2022nanosphere}, cracked template method \cite{zhu2021templateless}, nanoimprint lithography \cite{baracu2021silicon}, and direct writing \cite{li2022directly}. 
However, these mature technologies can still result in defects such as broken lines, metal deposition, photoresist residue due to lithography process parameters and operator errors.
In practical applications, metallic mesh can develop significant defects from scratches, impacts, and chemical erosion, which affect the performance of optoelectronic equipment.
Defect detection in transparent electromagnetic shielding films during production is often limited to manual visual inspection, with a lack of effective detection devices suitable for practical applications.
The development of a miniaturized, cost-effective portable device for microscopic defect detection could substantially improve the durability and performance of transparent electromagnetic shielding films.

Although the use of smartphone technologies in engineering detection has been limited, there has been notable progress in the development of portable diagnostic devices in this area. These smartphone-based devices offer notable advantages, including high performance, low cost, miniaturization, and portability. \cite{wang2023smartphone, bui2023smartphone}. Recent studies have documented a variety of applications for these microscopes, demonstrating their broad utility  \cite{huang2024deep, zheng2023hand, ahmed2023photonic}.
Smartphone microscopes are capable of identifying white blood cells \cite{janev2023smartphone}, imaging fresh tissue \cite{zhu2020smartphone}, and measuring environmental lead (Pb) content \cite{lai2022portable}.
The majority of these devices are designed as transmitted light microscopy systems, which are commonly used in biological and medical imaging. In this configuration, samples are positioned between the illumination and imaging systems, allowing light to pass through the sample.

The capacity to observe opaque objects is one of the significant extensions provided by the development of reflected light microscopy on smartphone-based platforms \cite{kim2023portable}.
%Zhu et al. have demonstrated a reflected microscopic system using a smartphone to flow cytometric fluorescent imaging \cite{zhu2011optofluidic}. LED light guided through the cross-section of the microfluidic device is transferred to samples through total internal reflection (TIR) and transferred to the imaging system. \Ming{what is the function of this following sentence? --> The LEDs are placed in the cross-section of the sample and are not integrated with the external optics.} Similarly, Wei et al. \cite{wei2017plasmonics} introduced a novel approach to smartphone-based fluorescence microscopy using surface-enhanced fluorescence (SEF) to significantly boost detection sensitivity. This technique involves inserting the samples directly into the microscope. Furthermore, Liu et al. have reported a microscopy with Ultraviolet Surface Excitation (MUSE) through the external lens and the frustrated TIR illumination of a 285 nm UV LED mounted on a smartphone for fluorescence imaging \cite{liu2021pocket}. However, the practical application of this technology is limited by a working distance of less than 1 mm and the need for precise focusing through the smartphone camera, which poses challenges for ultra-field-of-view imaging in industrial scenarios.
At the same time, most existing cell phone microscopes \cite{rabha2022affordable, kheireddine2019dual, lee2021smartphone} have a short working distance (less than 5 mm), making nondestructive detection during displacement scanning challenging. 
To address this limitation, we present a long working distance (22.23 mm) reflective smartphone microscope designed for the detection of defects in large micro- and nanostructures. This innovation paves the way for new applications in the industrial inspection field using smartphone microscopes.
%详细总结
%Design of a metal mesh grille reflective portable imaging system by combining a smartphone with an external optical component. The external optical components are rationally designed, integrated and packaged based on 3D printing technology, and coupled with the camera module of the smartphone to design a reflective portable imaging system to realize the function of microscopic imaging. The study analyzes the impact of the selected relay lens iPhone 7 front lens group on the imaging quality in forward and reverse positions. The stray light filtering design is introduced to study the imaging quality under different aperture diaphragm sizes, and the optimal aperture diaphragm value is given to effectively improve the imaging quality of the system.

On the software side, defect detection in structures such as metallic meshes remains challenging due to their complex textures and diverse defect characteristics.
% Metallic mesh, as a large-size micro-nanostructures, can be widely used in transparent electromagnetic shielding devices \cite{chung2023highly}, touch displays \cite{ouyang2022recent}, and transparent antennas \cite{yao2023transparent}. It features a complex texture and diverse defect characteristics. 
Typically, these precision devices undergo manual and visual inspections during the production stage, which lacks a systematic inspection model. Traditional methods like the spectral \cite{sakhare2015spectral} or statistical methods \cite{alper2014textural} often fall short in accuracy and efficiency, particularly in industrial settings where rapid and reliable assessments are crucial. Moreover, while deep learning approaches show promise, they require extensive datasets that are not always available \cite{jing2022mobile}.

Defect detection in periodic texture structures, such as fabrics, remains a focal research area within defect detection.  Researchers have developed several fabric defect detection algorithms utilizing approaches such as the spectral method \cite{sakhare2015spectral}, statistical method \cite{alper2014textural}, modeling method \cite{zhou2017fabric} and deep learning method \cite{jing2022mobile}. However, each method has its limitations. 
%The spectral method, often employed when defect contours are unclear, suffers from low detection accuracy, particularly in identifying small-area defects, due to the critical influence of filter parameter selection. The statistical method tends to overlook global information and can be computationally intensive, resulting in low accuracy and efficiency. Algorithms based on deep learning rely on a large number of datasets, and in many industrial fields, the dataset is an urgent problem, leaving the support of the dataset, the accuracy of the algorithm is difficult to ensure, and the algorithms of deep learning are not capable of accurately describing the shape of the defect contour. 
The spectral method struggles with accuracy for small defects due to filter sensitivity. The statistical method misses global information and is computationally heavy. Deep learning algorithms need extensive datasets, often lacking in industry, affecting accuracy. 
In recent years, defect detection algorithms based on low-rank decomposition models have gained significant attention in the modeling method, offering a promising direction for future research. These algorithms decompose images into low-rank (background), sparse (defect), and noise terms \cite{candes2011robust}. Ji et al. \cite{ji2020fabric} proposed WLRL to enhance defect-background separation using Laplace regularization, though it relies on limited Gabor features. Liu et al. \cite{liu2022fabric} introduced structural constraints to Robust Principal Component Analysis with noise term and defect prior (PN-RPCA), called SPN-RPCA, leveraging spatial connections and 8D texture features for defect highlighting. Bao et al. \cite{bao2022low} developed PTVLR, using SSIM as prior information. However, these methods are often inaccurate and time-consuming in extracting defect information in metallic mesh, leading to poor detection efficiency.

%This class of algorithms mainly decomposes the image into low-rank terms (background), sparse terms (defect) and noise terms \cite{candes2011robust}. Ji et al. \cite{ji2020fabric} have reported a new weighted low-rank decomposition model with Laplace regularization (WLRL) that can enlarge the distance between the background and the defective regions. This model requires dividing the periodic lattice and computing Gabor features within adjacent lattices, although these low-level features may not be comprehensive enough. Liu et al. \cite{liu2022fabric} proposed a low-rank decomposition model based on structural constraints (SPN-RPCA) incorporating structured sparsity paradigms that leverage the spatial connections of defective pixels. SPN-RPCA merges the energy image of the energy image of the fabric image with the original image to highlight the defective areas and then computes the 8D texture features as prior information to guide the RPCA. Bao et al. \cite{bao2022low} proposed a low-rank decomposition model (PTVLR) based on defective a prior information and a full-variance regular term. PTVLR introduces the structural similarity between the original fabric image and the reconstructed image (SSIM) as a prior information to guide the low-rank decomposition. These methods are inaccurate and time-consuming to extract the prior information about the defects of metallic mesh, which seriously affects the low-rank decomposition ability and leads to poor defect detection and low efficiency.

In this paper, we introduce a dual-prior weighted low-rank decomposition model (DW-RPCA) based on spectral filter fusion and the Hough transform to extract the prior information of defects in metallic mesh, guiding the accurate low-rank decomposition. 
Moreover, a robotic arm is used for image scanning of metal mesh grill over-field-of-view range detection. Combined with the miniaturized portable smartphone microscope, our system significantly improves the detection of defects in large-size micro- and nanostructures, contributing valuable advancements to the field of industrial inspection. This work represents an advancement in combining hardware development and algorithmic innovation to enhance the inspection capabilities for metallic mesh defects, paving the way for broader applications in industrial quality control.

\section{Methodology}
\subsection{Smartphone microscope for industrial defect detection}

\begin{figure}[!t]
	\centering
	\includegraphics[width=6.5in]{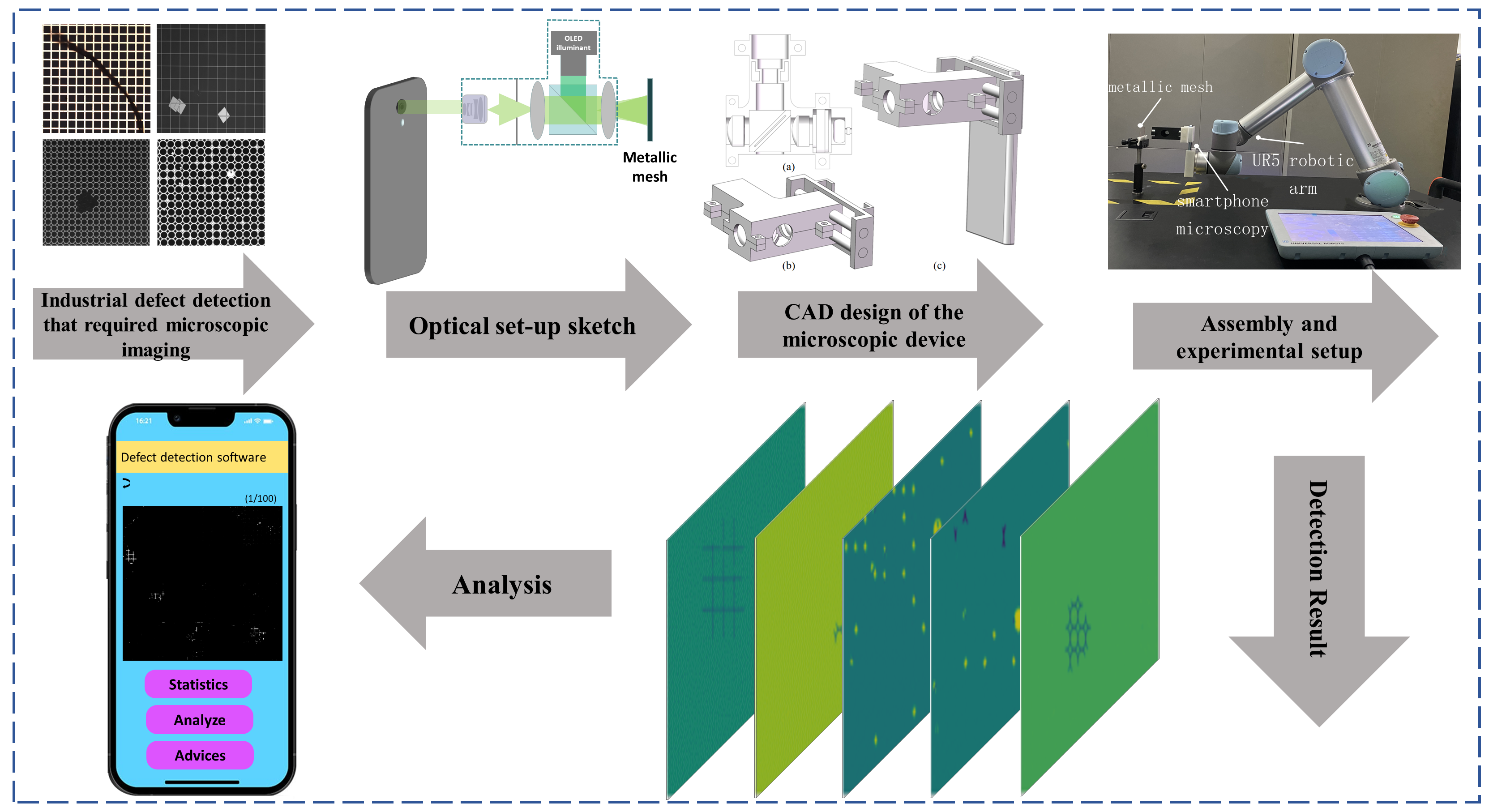}
	\caption{Schematic of the rapid prototyping of the LD-RSM system.}
	\label{fig_1}
\end{figure}

Fig. 1 presents the prototyping of the long-distance reflective smartphone microscope (LD-RSM) for industrial defect detection. 
This system is engineered to fulfill the requirements for real-time, in-situ detection of defects, featuring a reflective optical path that extends its working distance significantly.
The simplicity of its optical design not only ensures ease of production and deployment but also contributes to its robust performance. Key components of the LD-RSM include imaging optics, illumination optoelectronics, and a mechanical structure that attaches to the flange of a robotic arm, all encased within a durable 3D-printed housing.
The optical design of the present imaging system is depicted in Fig. 2(a) and has been discussed in detail in Section 2.2.
A monochromatic high-power coaxial point light source with a wavelength of 550 nm is employed as the illumination source for LD-RSM, thereby mitigating the impact of chromatic aberration on the imaging quality of the system.
Components are securely mounted inside a housing produced by a fused-deposition modeling (FDM) 3D printer, model Raise N2.
Industrial samples are held in place by an optical clamping device, ensuring stability during inspections. 
The three-dimensional (3D) rendering of the LD-RSM system is presented in Fig. 2(b), and the prototype of the developed microscopic system is showcased in Fig. 2(c).

\subsection{Imaging system design}

\begin{figure}[!t]
	\centering
	\includegraphics[width=6.5in]{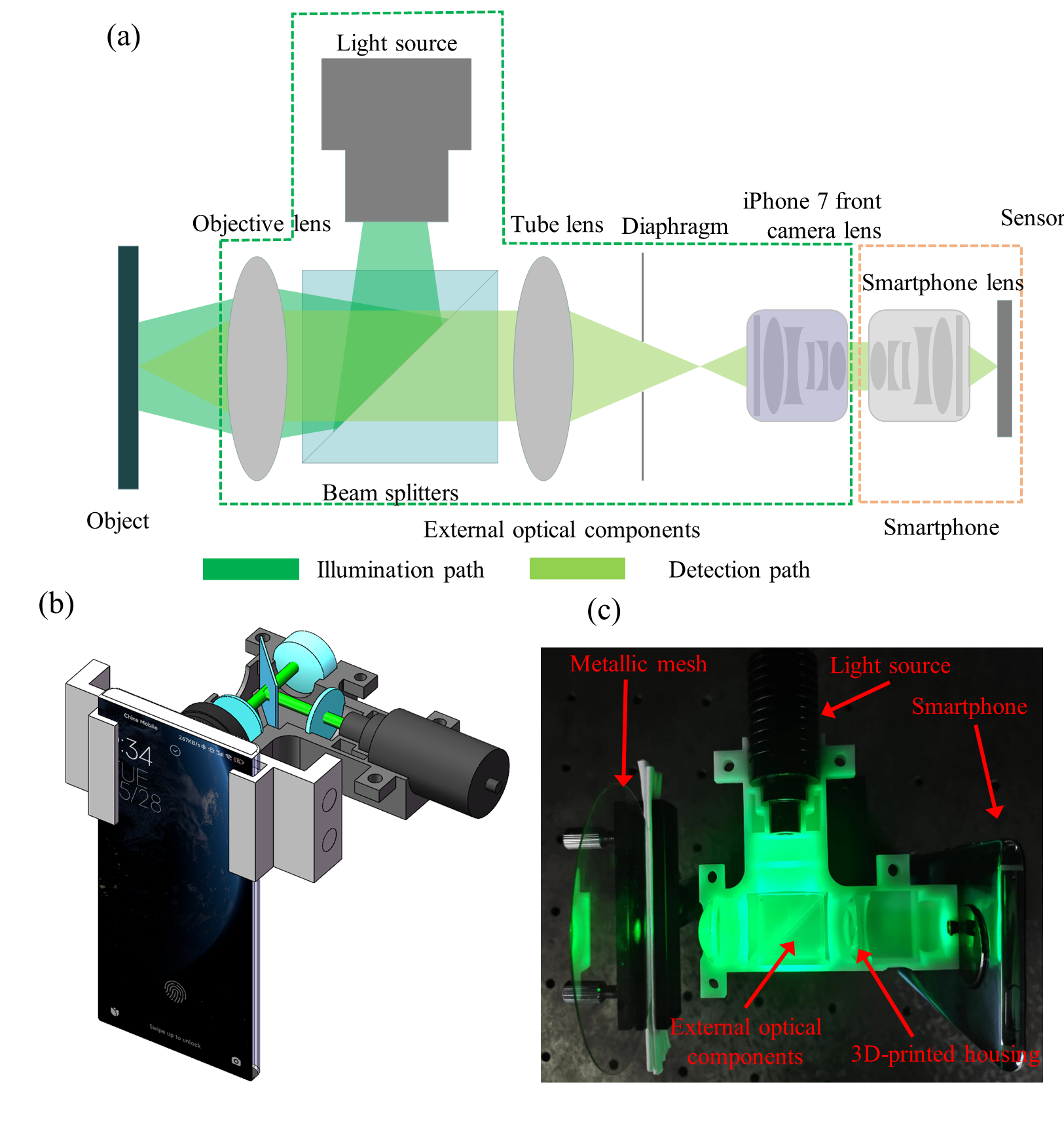}
	\caption{Design and fabrication of the LD-RSM system. (a) Schematic of a 4\textit{f} optical configuration.  (b) 3D rendering of the LD-RSM system and (c) Prototype of the developed microscopic system.}
	\label{fig_2}
\end{figure}

There are two conventional approaches to configuring a microscopic setup on a smartphone: the 3\textit{f} and 4\textit{f} imaging systems \cite{rabha2022programmable}. 
In 3\textit{f} configuration, an external lens is mounted directly to the outside of the camera module of the smartphone to form a finite conjugate system.
The external lens works as the objective lens and the built-in lens set of the smartphone works as the tube lens of the imaging system. 
The sample is positioned at the focal plane of the objective lens. The optical field after the sample passes through the objective and tube lenses, then  captured on the smartphone sensor. 
Given the short focal length of the smartphone lens (typically only a few millimeters), the external lens must have a similar short focal length to achieve greater magnification for microscopic imaging.
%This article also verifies this viewpoint through experiments.
However, due to its small magnification capability and the constraints imposed by the pixel pitch of the smartphone sensor, the setup often struggles with limited resolution and field of view, making it less suitable for practical industrial inspection.

To develop a high-quality, cost-effective optical system, we opted for the 4\textit{f} imaging configuration for our LD-RSM configuration. 
This setup includes two long focal length, large diameter lens as the objective and tube lens,
and incorporates a beam splitter, diaphragm, and light source, making it ideal for in-situ imaging measurements of metallic mesh. The imaging principle is shown in Fig. 2(a).
Both the illumination and imaging optical paths are aligned on the same side of the sample to enable imaging measurements during equipment servicing without disassembly.
Light from the source first passes through a beam splitter, which directs the reflected light onto the sample via the objective lens. The sample is placed at the focal plane of the objective lens. The reflected light carrying sample information then passes through the objective lens and the tube lens, forming an image on the focal plane of the tube lens. This image is then transferred to the smartphone camera module through a relay lens and captured on the camera sensor.
%Light from the source first passes through a beam splitter, with the reflected light focusing onto the sample via the objective lens. The reflected light, now carrying sample information, goes through the objective and tube lenses to be focused by the tube lens. It is then relayed through iPhone 7 front camera lens to the sensor of the smartphone camera module, ensuring detailed imaging of the sample. 
%\Ming{[''ensuring detailed imaging of the sample'' It is not clear explaining like this...]}

The imaging system employs a Redmi K30 Ultra smartphone (Xiaomi Inc.) equipped with a Sony IMX582 CMOS sensor (64-megapixel with pixel size of 0.8 $\mu$m). The built-in lens of the smartphone features a focal length of 5.43 mm. 
Since the imaging focal length of the smartphone camera module is only 5.43 mm, the focal length of the relay lens cannot be too large considering the magnification of the system. The relay lens utilized here is derived from the iPhone 7 front camera module, which offers an imaging focal length of 2.87 mm and a numerical aperture (NA) of 0.23.  Compared with the widely used ball lens and achromatic objective lens, a better imaging effect can be obtained by using the cell phone lens set.

 LD-RSM system requires placing the sample at the focal plane of the objective lens, where the focal length of the objective lens is crucial for determining the working distance. %Meanwhile, considering the influence of ambient light on the imaging quality, the use of achromatic double glued lens as the objective lens can further reduce the influence of chromatic aberration and improve the imaging quality of the system. 
In this paper, the objective lens is an achromatic double-glued lens from Edmund, with a focal length of 30 mm, a diameter of 25 mm, and a numerical aperture NA of 0.38. A long working distance of 22.23 mm is realized.
The tube lens is integral for converging imaging to achieve adequate magnification while maintaining a compact form factor for the external device. We use an Edmund plano-convex lens with a focal length of 40 mm and a diameter of 25 mm.
LD-RSM utilizes light from the transmitted portion of the beamsplitter to image the sample, and the reflected portion is used for sample illumination. The Edmund beamsplitter has a size of 35 mm × 35 mm and a reflectance to transmittance ratio of 50/50. %At the same time, in order to reduce the impact of chromatic aberration on the imaging quality of the system, 
We use a monochromatic high-power coaxial point light source with a wavelength of 550 nm.

\begin{figure}[!t]
	\centering
	\includegraphics[width=5in]{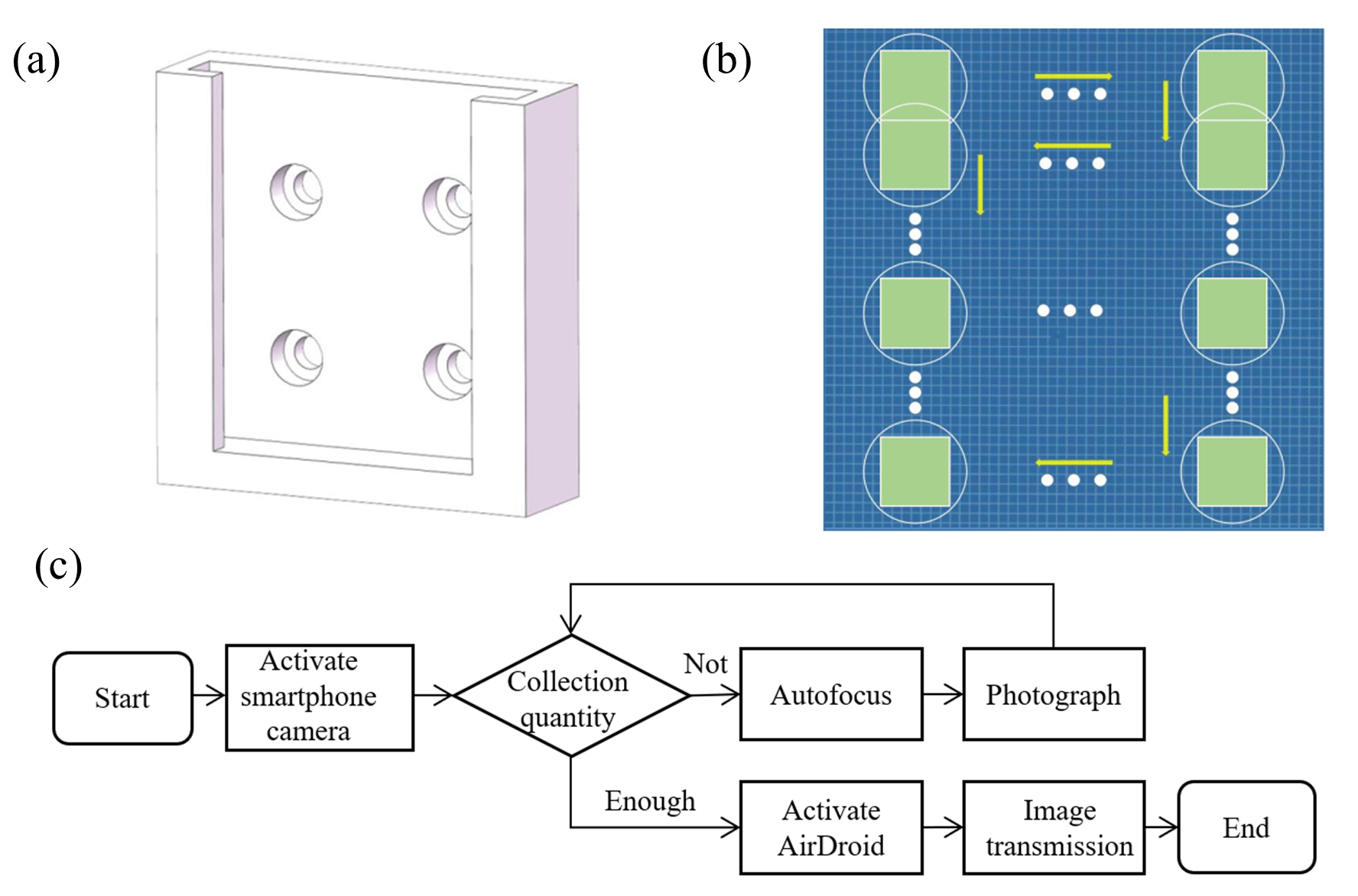}
	\caption{Image acquisition and transmission strategies. (a) Fixed mechanics for UR5 arm and LD-RSM, (b) UR5 robotic arm scanning programme, (c) LD-RSM Script for automatic image acquisition and transmission.}
	\label{fig_3}
\end{figure}

\subsection{Imaging acquisition and transmission}

%Metallic mesh  are used as a large-size micro-nano structure with micrometer line widths and sub-millimeter periods. At the same time the overall size reaches the decimeter scale. Therefore, a precision displacement device equipped with the LD-RSM is required to realize the detection of its defects. 
In this paper, the UR5 robotic arm was chosen to be programmed to scan and acquire images of metal grids within a specific area.
The tool flange of the UR5 arm is equipped with four M6 threaded holes which can be used to connect the UR5 arm to the LD-RSM. According to the parameters of the tool output flange of the UR5 arm, the mechanical structure in Fig. 3(a) is designed by Solidworks software, which is used to connect the LD-RSM, and serves as a fixed support.
The scanning scheme is designed as shown in Fig. 3(b). The scheme is programmed through the PolyScope to control the UR5 robotic arm, which carries the LD-RSM and automatically scans the area to be inspected of the metallic mesh in an S-path to obtain the image of the area. 
In order to ensure that defects missed detection due to the motion error of the arm, a certain amount of redundancy between the images is guaranteed. 
The UR5 arm moves 500 $\mathrm{\mu}$m, design a node, UR5 in each node of the residence time is set to 2s, to ensure that the LD-RSM can focus on the image, and complete the image acquisition operation.

The control flow of the script program is shown in Fig 3(c). The automatic acquisition begins with activating the camera, then the number of images to be captured is judged, and the focusing and photo-taking operations are performed cyclically until the required number is met, finally the AirDroid software is turned on to transmit the images. %This method has the advantages of automatic image acquisition and transmission, simple operation and high efficiency. 

\begin{figure}[!t]
	\centering
	\includegraphics[width=6.5in]{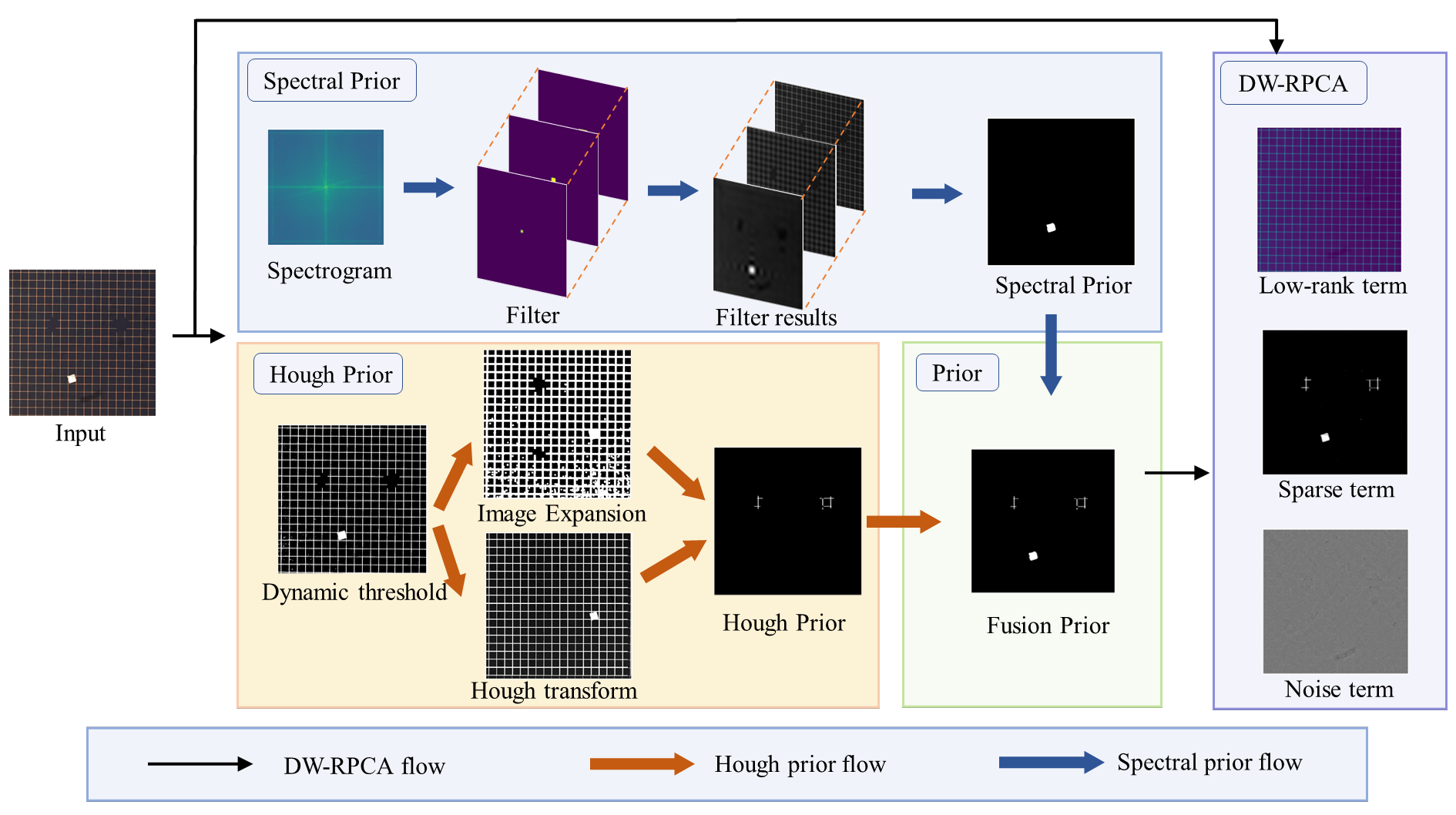}
	\caption{The overall detection process of the DW-RPCA.}
	\label{fig_4}
\end{figure}

\subsection{Low-rank decomposition defect detection based on spectral filter fusion and Hough transform}

The defects that affect the photoelectric performance of metallic mesh fall into two categories: broken line and block defects. 
Based on the characteristics of these two types of defects, we propose a dual prior weighted Robust Principal Component Analysis (DW-RPCA) for metallic mesh defect detection.
The overall detection process of the DW-RPCA proposed is shown in Fig.4, which is mainly composed of three parts: prior information extraction, RPCA and threshold segmentation. 

In the prior information extraction phase, spectral filter fusion prior and Hough transform prior are designed for defects of square metallic mesh and circular metallic mesh, respectively. These priors guide the low-rank decomposition process.
As shown in Fig.4, we use Fourier transform to get the spectrogram whose frequency gradually increases from the center to the periphery. In this paper, three square low-pass filters $\mathit{f}_1$, $\mathit{f}_2$ and $\mathit{f}_3$ are designed, and the side lengths of the square low-pass filters are 10, 20 and 40 respectively. These filters are applied to the spectrogram obtained before. The Fourier inverse transform is then performed on the filtered spectrogram to obtain $\mathit{I}_1$, $\mathit{I}_2$ and $\mathit{I}_3$.

% \begin{equation}
% F(u, v)=\frac{1}{M N} \sum_{x=0}^{M-1} \sum_{y=0}^{N-1} f(x, y) e^{-j 2 \pi(2 x / M+w / N)}
% \end{equation}

% \begin{equation}
% f(x, y)=\frac{1}{M N} \sum_{u=0}^{M-1} \sum_{v=0}^{N-1} F(u, v) e^{j 2 \pi((2 c / M+y / N)}
% \end{equation}

We utilize a linear combination to fuse the images obtained after filtering with three different sizes of filters—large, medium, and small, respectively:
\begin{equation}
P=k_1 \times I_3+k_2 \times\left(I_1+I_2\right)-k_3 \times\left(I_2-I_1\right),
\end{equation}
where $\mathit{k}_1$, $\mathit{k}_2$ and $\mathit{k}_3$ are 0.8, 0.2 and 0.3.

Finally, the prior information about block defects is obtained by binarizing the resulting fusion map.
The Hough transform and the Hough circular transform are invoked to construct the broken line defect prior for square metallic mesh and circular metallic mesh, respectively.
As shown in Fig. 4, firstly, the input image is binarised. Then, the resultant map is created by detecting straight lines using the Hough transform and plotting these lines on the input image. Simultaneously, the resultant map undergoes an expansion operation. Finally, the two resultant maps are combined. The final map is then binarized to serve as the broken line defect prior.

In DW-RPCA, the prior information is used to guide the decomposition process in the sparse part, ensuring that defects that do not satisfy sparsity are also decomposed into the sparse part. The model is as follows:

\begin{equation}
\begin{aligned}
& \min _{A, E}\|A\|_*+\lambda\|W \odot E\|_1+\frac{\beta}{2}\|N\|_F^2 , \quad \text { s.t. } D=A+E+N,
\end{aligned}
\end{equation}
where  ${{\mathit{D}}}$, $ {{\mathit{A}}}$, ${{\mathit{E}}}$, ${{\mathit{N}}}$ $\in {{\textit{R}}^{\textit{d} \times \textit{d}}}$ are the input image, background matrix, defect matrix, noise matrix, respectively. 
${{\mathit{W}}} \in {{\textit{R}}^{\textit{d} \times \textit{d}}}$ is the weight matrix.
 $\left\| {{\mathit{A}}} \right\|_*=\text{Tr}(({{\mathit{A}}^\textit{T} {\mathit{A}}})^\text{1/2})$ denotes the nuclear norm of the low-rank matrix ${\mathit{A}}$.
$\left\| {{\mathit{E}}} \right\|_1$ represents the 1-norm of the sparse matrix ${\mathit{E}}$. $\left\| {{\mathit{N}}} \right\|_{\mathrm{\textit{F}}}^\text{2}$ is the Frobenius norm of the noise term ${{\mathit{N}}}$.
$'\odot'$ is the Hadamard product, and the parameters $\mathit{\lambda}$ and $\mathit{\beta} $ are weights of different terms. 

The Lagrangian function of the model is constructed as follows:
\begin{equation}
\begin{aligned}
& \min _{A, E, N, Y, \mu} L(A, E, N, Y, \mu) \\
& =\|A\|_*+\lambda\|W \odot E\|_1+\frac{\beta}{2}\|N\|_F^2+  \langle Y, D-A-E-N\rangle+\frac{\mu}{2}\|D-A-E-N\|_F^2
\end{aligned}
\end{equation}

Using the  alternating direction method of multipliers (ADMM) \cite{lin2011linearized} to solve the above equation, the specific solution procedure after introducing the prior information and adding the weight matrix is as follows:
\begin{equation}
E_{k+1}=\arg \min L\left(A_k, E, Y_k, \mu_k\right)=\underset{E}{\arg \min } \frac{\lambda}{\mu_k}\|W \odot E\|_1+\frac{1}{2}\left\|E-X_E\right\|_F^2
\end{equation}
where $X_E=D-A_k-N_k+Y_k / \mu_k$. The L1-norm optimization problem with weight matrices can be transformed into an operation between individual elements:
\begin{equation}
\begin{aligned}
E_{k+1}(i, j) & =S_{\varepsilon(i, j)}(X(i, j)) =\operatorname{sgn}(X(i, j)) \times \max [\operatorname{abs}(X(i, j))-\varepsilon(i, j), 0]
\end{aligned}
\end{equation}
where $\varepsilon(i, j)=\lambda W(i, j) / \mu_k$, ${S_{\varepsilon}}$ denotes the soft-threshold operator, a solving procedure that replaces the dot-multiplication of matrices with operations between single elements.
The overall flow of the DW-RPCA is shown in Table 1.

\begin{table}[width=.9\linewidth,cols=4,pos=h]
\caption{Alternate Direction Method (ADM) flow chart for solving the DW-RPCA model.}\label{tbl1}
\begin{tabular*}{\tblwidth}{@{} LLLL@{} }
\toprule
Concrete content \\
\midrule
{\textbf{Input:}} Metallic mesh  image ${{\mathit{I}}}$, defect prior  ${{{W}}} $, parameter  ${\alpha}, {\beta} $, $\textit{p} = {{0.75}}$, $ \tau ={{30}} $;   \\
 {\textbf{Initialize:}}  ${{{\mathit{L}}}}={{{\mathit{E}}}}={{{\mathit{N}}}}={\textit{u}}= {\text{0}}$, $\rho  = {\text{0.8}} $, ${\textit{maxstep}}= {\text{10}} $,  $\varepsilon  = {\text{1}\textit{e}-\text{4}} $, $\textit{k} = {\text{0}} $ \\
 {\textbf{while} not converged or $\textit{k}<{\textit{maxstep}}$ \textbf{do}} \\
1. Update  ${{\mathit{L}}} $ by \\
 $  \begin{array}{l}
	    	{{{\mathit{L}}}} = \arg \mathop {\min }\limits_{{\mathit{L}}} {\rm{ }}{\left\| {{{{\mathit{L}}}}} \right\|_{\textit{Sp},\tau }}{\rm{ + }}\frac{\rho}{\text{2}}\left\| {{{{\mathit{L}}}} -{{{\mathit{I}}}} +{{{\mathit{E}}}} +{{{\mathit{N}}}} - {\textit{u}}} \right\|^\text{2}_{\mathrm{\textit{F}}}
	    	\end{array} $ \\
2. Update  ${{\mathit{E}}} $ by	 \\
$\begin{array}{l}
    	{{{\mathit{E}}}} = \arg \mathop {{\rm{min}}}\limits_{{\mathit{E}}} {\rm{ }}{\alpha}{\| {{\mathit{W}}}\odot{{\mathit{E}}} \|_\text{1}}+\frac{\rho }{\text{2}}\left\| {{{{\mathit{E}}}} - {{\mathit{A}}}} \right\|^\text{2}_{\mathrm{\textit{F}}}\\
    	{\rm{where }} \enspace {{ {\mathit{A}}}} = {{\mathit{I}}} - {{{\mathit{L}}}} - {{{\mathit{N}}}} + {{{\textit{u} }}}
       \end{array} $ \\
3. Update  ${{\mathit{N}}} $ by	\\
$  \begin{array}{l}
       {{{\mathit{N}}}} = \frac{\rho }{\text{2}\mathit{\beta} + \rho} ({{\mathit{I}}} - {{{\mathit{L}}}} - {{{\mathit{E}}}} + {{\textit{u} }}) 
        \end{array} $ \\
4. Update  ${\textit{u}} $ by	\\
$  \begin{array}{l}
		{\textit{u}} = {\textit{u}} + {{\mathit{I}}} - {{{\mathit{L}}}} - {{{\mathit{E}}}} - {{{\mathit{N}}}}
		\end{array} $ \\
5. Check the convergence conditions
		$  \begin{array}{l}
	 {\left\| {{{\mathit{I}}} - {{{\mathit{L}}}} - {{{\mathit{E}}}} - {{{\mathit{N}}}}} \right\|_\infty } < \varepsilon \enspace {\rm{ and }} \enspace \textit{k}{\rm{ <  {\textit{maxstep}}}}
		\end{array} $	\\
6. Update  $\textit{k} = {\textit{k+}\text{1}} $ \\
{\textbf{end while}}   \\

\bottomrule
\end{tabular*}
\end{table}

After the DW-RPCA, defects need to be extracted by threshold segmentation.
Since the gray value of broken line defects in the sparse part is less than zero, while the gray value of block defects is greater than zero, and the absolute value of the gray value corresponding to broken line defects is small, while the absolute value of the gray value corresponding to block defects is large. Therefore, the defects are extracted using double threshold segmentation, as shown in equation (6) 

\begin{equation}
E(x, y)= \begin{cases}255, & \text { if } E(x, y) \leq t_1 \\ 255, & \text { if } E(x, y)>t_2 \\\ 0, & \text { else }\end{cases}
\end{equation}

%The threshold value $t_1$ is -5 and $t_2$ is 70, and the sparse part corresponding to the square mesh grid is double threshold segmentation, and the segmentation knot process is shown in Fig. 16, and the double threshold segmentation is able to effectively extract the defects of the metal mesh grid in the sparse part. The threshold $t_1$ corresponding to the circular mesh grating is -30, and the threshold $t_2$ is 50.

\subsection{Sample preparation}
The line widths of the metallic mesh are on the micron scale and are therefore processed using UV lithography, metal sputtering and stripping processes \cite{wang2019highly}.
Specifically, mask mapping was first carried out using L-Edit software, and then mask fabrication was carried out using electron beam direct writing equipment. 
This is followed by substrate pre-treatment, spin-coating of photoresist, pre-baking, UV exposure, post-baking, developing and rinsing and drying.
After the above process, the metallic mesh pattern on the mask plate can be successfully transferred to the substrate photoresist.
It should be noted that the negative photoresist type AR-N 4400-05 from ALLRESIST was selected for the subsequent stripping process, which has the advantage of very high sensitivity and ease of removal, as well as very high resolution of the contours of the photoresist pattern after development.
Next, the metal layer was deposited on the adhesive film samples using a Technocor JCPF-2000A ultra-high vacuum dual target magnetron sputtering machine, and the final thickness of the deposited metal layer was about 100 nm.
The final preparation of the metallic mesh was stripped by immersing it in acetone solution combined with low power ultrasonication and brush scrubbing, and washed and dried.

\section{Results and discussion}

\subsection{Imaging system characterization}
To obtain the best results from the LD-RSM system, an optimal distance of 22.23 mm has been maintained between the sample and the external lens assembly edge. 
The optical magnification of LD-RSM is M1 (the ratio of the tube lens FL = 40 mm to the objective lens FL = 30 mm) times M2 (the ratio of the phone’s internal lens FL = 5.43 mm to the relay lens FL = 2.87 mm) = Mo = 2.52x.
The phone has a screen size of 6.67 inches (diagonally) with an aspect ratio of 20:9, and its imaging sensor size is 1/1.7 inches (diagonally).
Thus, the displayed image will have a digital magnification of  Md = 18.0x on the phone’s screen.

The resolution of  LD-RSM has been characterized by imaging 1951 USAF target. This is shown in Fig. 5 (a). 
Experimentally, under green light illumination, we measure a resolution of 4.92 $\mu$m, as it can resolve element 5 in group 6.

Due to the roughness of the sample surface, along with scattering and reflection within the optical system, the LD-RSM has significant stray light interference, compared to transmission microimaging systems.

The presence of stray light in the LS-RSM system primarily arises from three sources. First, the uneven illumination from the LED source contributes to non-uniform light distribution. Second, scattering is induced by irregularities in the surface structure of the metal mesh grid. Meanwhile,, stray light is also caused by reflections from multiple optical elements and the inner wall of the housing.

The diaphragm is situated in the focal plane to the left (near the tube lens). 
Note that the relay lens, adapted from an iPhone 7 front lens with a short 2.87 mm focal length, complicates precise diaphragm placement. Despite slight positional inaccuracies, this arrangement successfully reduces stray light and enhances imaging quality. 
To avoid degradation of image clarity from light reflecting off the shell and re-entering the imaging path, an opening is strategically placed in the transmitted path after the splitter. 
A comparison of the imaging results before the improvement and after the measures to suppress stray light is shown in Fig. 5(b) and (c). The imaging results show that the impact of stray light on the imaging quality of the system has been greatly reduced by the above improvement measures, and the imaging clarity has been significantly improved.

\begin{figure}[!t]
	\centering
	\includegraphics[width=3in]{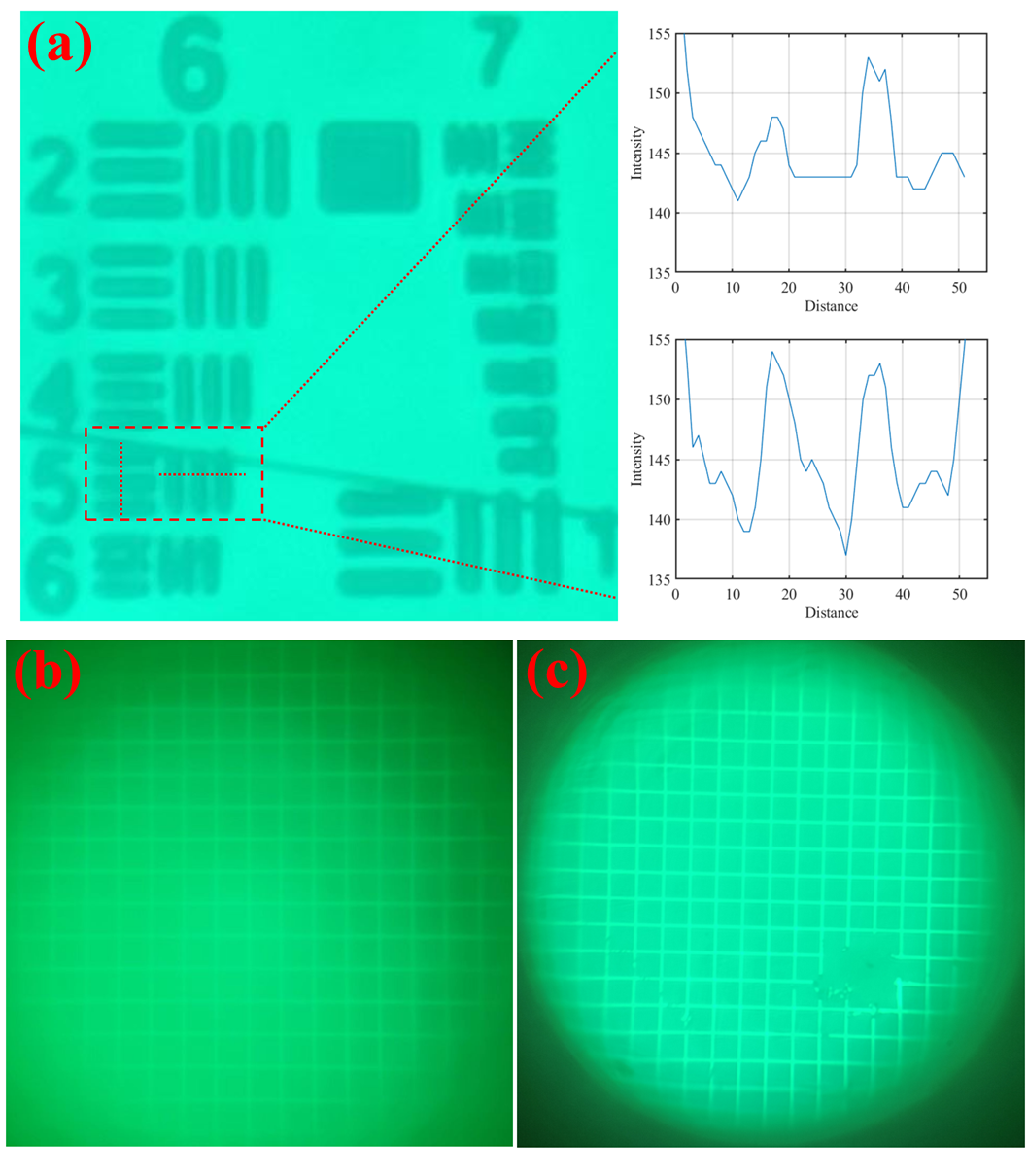}
	\caption{Characterization of the LD-RSM system. (a) Image
of a USAF-1951 resolution test target acquired by the LD-RSM. (b) Imaging result before improvement. (c) Imaging result after the measures to suppress stray light.}
	\label{fig_5}
\end{figure}

\subsection{Quantitative analysis of DW-RPCA by metallic mesh defect detection}

\begin{figure}[!t]
	\centering
	\includegraphics[width=5in]{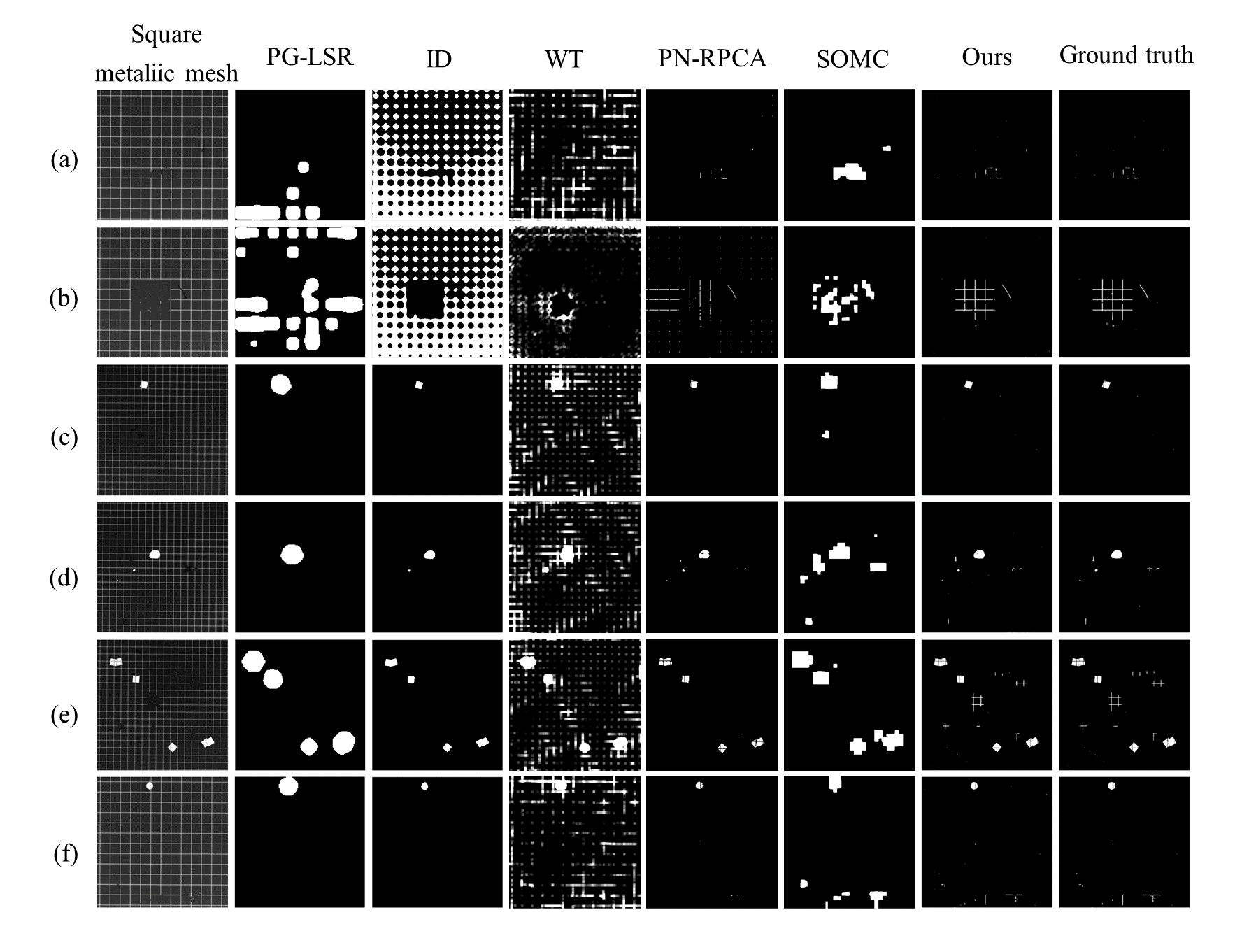}
	\caption{Visual comparisons with four representative methods on square metallic mesh database.}
	\label{fig_6}
\end{figure}

Two types of defects mainly affect the photoelectric performance of metallic mesh: broken lines and lumpy defects. To verify the performance of the model DW-RPCA, this paper prepares the metallic mesh of square and circular structures and utilizes the laser confocal microscope OLYMPUS OLS5000 to prepare a set of 60 images containing broken lines, lumpy, and mixed defects. Labelme software \cite{russell2008labelme} was used to label the ground truth.

In order to quantitatively assess the state-of-the-art of the model, the True Positive (TP), True Negative (TN), False Positive (FP), and False Negative (FN) are introduced for the detection performance of the defect detection algorithms. From these values, we derive the following as evaluation metrics which are critical for performance assessment:

\begin{equation}
{\text{TPR} = }\frac{{{\text{TP}}}}{{\text{FN}}+{\text{TP}}}, 
\quad	{\text{FPR} = }\frac{\text{FP}}{{\text{FP}}+{\text{TN}}},
\quad	{\text{PPV} = }\frac{{{\text{TP}}}}{{\text{FP}}+{\text{TP}}},
\quad	{\text{NPV} = }\frac{{{\text{TN}}}}{{\text{TN}}+{\text{FN}}},
\end{equation}

\begin{equation}
f=\frac{\left(\gamma^2+1\right) \times TPR \times PPV}{TPR+\gamma^2 \times PPV},
\end{equation}
where the value of $f$ combines the true rate and the positive predictive value, and we set $\gamma$ = 1.

\begin{table}[htbp]
\centering
\caption{Quantitative comparisons on square metallic mesh datasets, the best results are marked with bold.}
\begin{tabular}{cccccccl}
\toprule
Defect Type & TPR & FPR & PPV & NPV & \( f \) & Method \\
\midrule
\multirow{4}{*}{Broken} & 0.3909 & 0.1565 & 0.0086 & 0.9966 & 0.0168 & PG-LSR \\
                       & 0.1821 & 0.4680 & 0.0009 & 0.9939 & 0.0019 & ID \\
                       & 0.3413 & 0.2579 & 0.0046 & 0.9956 & 0.0091 & WT \\
                       & 0.5097 & 0.0011 & 0.7423 & 0.9974 & 0.6009 & PN-RPCA \\
                       & 0.6905 & 0.0298 & 0.0730 & 0.9980 & 0.1274 & SOMC \\
                       & \textbf{0.8158} & \textbf{0.0003} & \textbf{0.8860} & \textbf{0.9995} & \textbf{0.8480} & Ours \\
\\
\multirow{4}{*}{Block} 
                       & 0.9227 & 0.0237 & 0.1372 & \textbf{0.9997} &  0.2356 & PG-LSR \\
                       & 0.8667 & 0.0445 & 0.8148 & 0.9994 & 0.8023 & ID \\
                       & 0.9456 & 0.2592 & 0.0172 & \textbf{0.9997} & 0.0334 & WT \\
                       & 0.7768 & \textbf{0.0001} & \textbf{0.9892} & 0.9989 & 0.8651 & PN-RPCA \\
                       & \textbf{0.9713} & 0.0252 & 0.1429 & 0.9995 & 0.2464 & SOMC\\
                       & 0.8772 & 0.0004 & 0.8769 & 0.9995 & \textbf{0.8753} & Ours \\
\\
\multirow{4}{*}{Mixed} & 0.5720 & 0.0365 & 0.1142 & 0.9972 & 0.1886 & PG-LSR \\
                       & 0.5388 & 0.0918 & 0.7114 & 0.9963 & 0.5783 & ID \\
                       & 0.5750 & 0.2441 & 0.0218 & 0.9950 & 0.0412 & WT \\
                       & 0.3667 & \textbf{0.0001} & \textbf{0.9841} & 0.9943 & 0.5174 & PN-RPCA \\
                       & \textbf{0.8605} & 0.0441 & 0.1423 & 0.9986 & 0.2414 &  SOMC \\
                       & 0.8157 & 0.0006 & 0.8815 & \textbf{0.9989} & \textbf{0.8460} & Ours \\
\bottomrule
\end{tabular}
\end{table}

In the DW-RPCA model, there are 2 weighting parameters $\mathit{\lambda}$ and $\mathit{\beta} $. Using the gridded parameter scanning method, $\mathit{\lambda}$ = 0.11, $\mathit{\beta} $ = 0.003 were determined in the square metallic mesh image set. We set $\mathit{\lambda}$ = 0.06, $\mathit{\beta} $ = 0.004 in the circular metallic mesh.
   
To validate the performance of DW-RPCA, we conducted comparisons with traditional defect detection algorithms including PG-LSR \cite{cao2017fabric}, ID \cite{ng2014patterned}, WT \cite{46-WT}, PN-RPCA \cite{candes2011robust}, SOMC \cite{14fabricSOMC}. 
PG-LSR is formulated as a least squares regression-based subspace segmentation model that incorporates a simple and effective prior learned from the image's local texture features.
ID posits that a defective fabric image comprises a superposition of defective objects (cartoon structure) and patterned fabric (texture structure). To regularize these structures, total variation and semi-norm in negative Sobolev space are employed for the cartoon and texture components, respectively.
WT utilizes low-level features from the wavelet transform domain. Significant maps are generated by calculating local center-periphery differences and global contrast using these features.
PN-RPCA incorporates a noise term and defect prior based on Robust Principal Component Analysis (RPCA), identifying sparse terms as defects.
Utilizing second-order gradient information, which better characterizes fabric texture features, SOMC designed a novel second-order multi-channel feature extraction method. This method simulates the response and distribution characteristics of P-type ganglion cells in the primate retina.

Validation is conducted using the square metallic mesh defect image dataset, with representative results shown in Fig. 6. Quantitative comparisons of different methods' measurement metrics are listed in Table 2.
PG-LSR and WT struggle to detect broken line defects. PG-LSR integrates local priors with the global feature space structure, making it sensitive to block defects. Although it has a low false detection rate, it fails to identify smaller defects.
WT has high FPR and low PPV due to its ability to detect block defects, but with significant noise.
Similarly, ID fails to detect broken line defects and may misidentify other elements.
PN-RPCA can basically detect broken wires and block defects, but it fails to detect mixed type defects, resulting in significant missed detections. Although FPR is low, TPR is poor.
SOMC has a high TPR but a low F-number. This is because SOMC can detect defects, but the significant blocks are large, and it also misses mixed defects.
In contrast, DW-RPCA demonstrates a clear advantage in detecting defects in square metallic mesh data.

\begin{figure}[!t]
	\centering
	\includegraphics[width=5in]{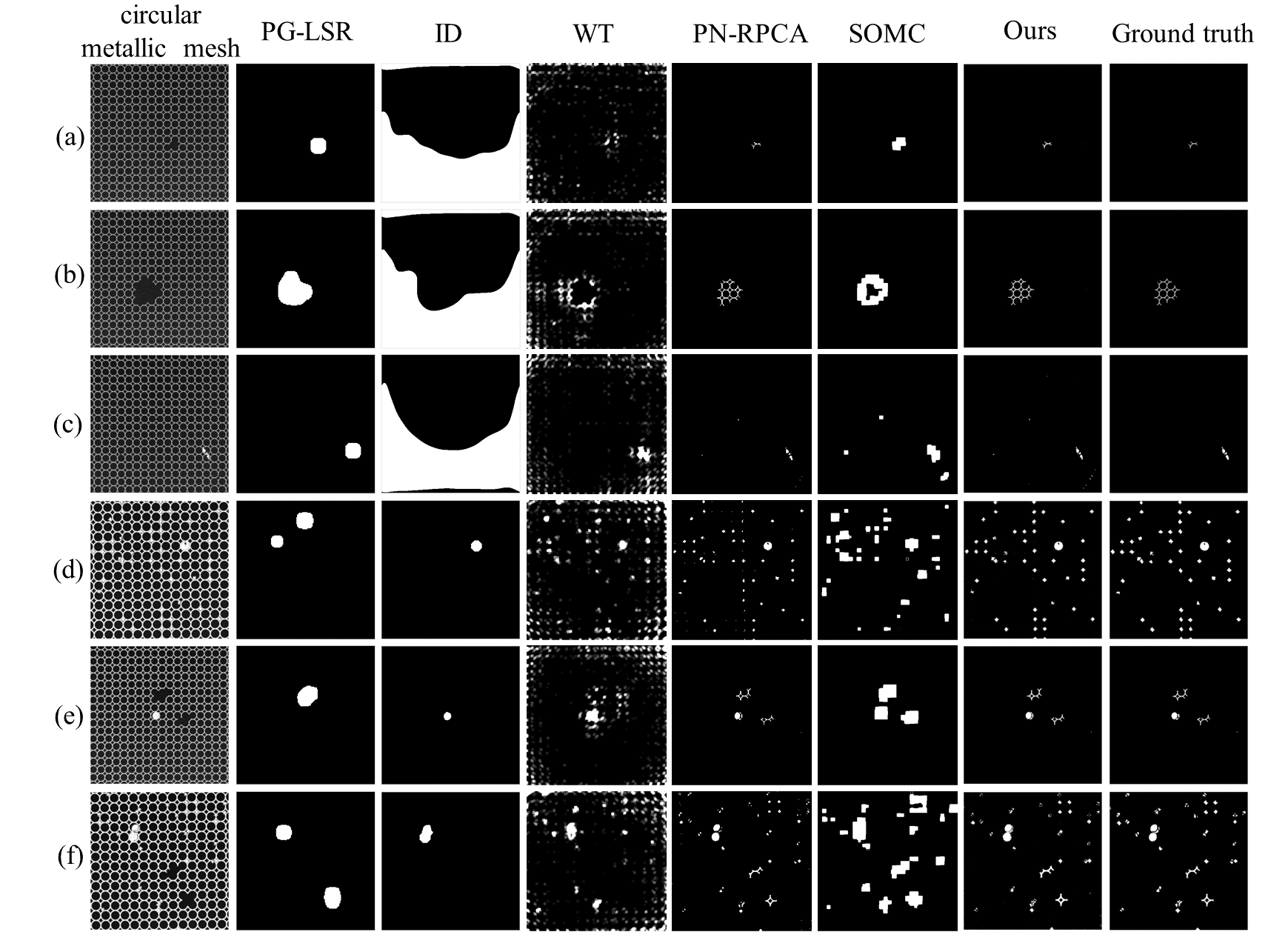}
	\caption{Visual comparisons with four baseline methods on circular metallic mesh database.}
	\label{fig_7}
\end{figure}

Fig. 7 presents comparisons of the circular metallic mesh database. 
PG-LSR can generally locate significant defects, however, defect characterization is difficult to achieve the classification of defects. 
ID fails to detect broken line defects and small-sized block defects.
WT can detect obvious defects, but it generates significant false positives due to noise.
The inability of the 8D texture method in PN-RPCA to extract prior information about small-area block defects results in poor performance in detecting these defects.
SOMC can detect most defects but struggles to distinguish between defect types.
DW-RPCA can accurately localize and characterize defects, closely matching the ground truth.
Quantitative comparisons of different methods' measurement metrics are listed in Table 3.
It can be seen that PN-RPCA has advantages in FPR and PPV, but TPR and f-number are not as good as DW-RPCA, because the former has more missed detections.
SOMC has a higher TPR but a lower PPV, while DW-RPCA offers comprehensive advantages across metrics.

\begin{table}[htbp]
\centering
\caption{Quantitative comparisons on circular metallic mesh datasets, the best results are marked with bold.}
\begin{tabular}{cccccccl}
\toprule
Defect Type & TPR & FPR & PPV & NPV & \( f \) & Method \\
\midrule
\multirow{5}{*}{Broken} & 0.6159 & 0.0246 & 0.0539 & 0.9996 & 0.0983 & PG-LSR \\
                       & 0.1821 & 0.4680 & 0.0009 & 0.9939 & 0.0019 & ID \\
                       & 0.2466 & 0.1985 & 0.0049 & 0.9980 & 0.0096 & WT \\
                       & 0.4469 & \textbf{0.0001} & \textbf{0.9382} & 0.9975 & 0.6035 & PN-RPCA \\
                       & 0.9008 & 0.0219 & 0.1508 & 0.9993 & 0.2537 & SOMC \\
                       & \textbf{0.9564} & 0.0004 & 0.8089 & \textbf{0.9998} & \textbf{0.8716} & Ours \\
\\
\multirow{5}{*}{Block} & 0.3350 & 0.0333 & 0.1110 & 0.9908 & 0.1360 & PG-LSR \\
                       & 0.8667 & 0.0445 & 0.8148 & \textbf{0.9994} & 0.8023 & ID \\
                       & \textbf{0.9629} & 0.2131  & 0.0512 & 0.9993 & 0.0948 & WT \\
                       & 0.6079 & 0.0066  & \textbf{0.8365} & 0.9946 & 0.6566 & PN-RPCA \\
                       & 0.7352 & 0.0414 & 0.1496 & 0.9958 & 0.2446 &  SOMC \\
                       & 0.8479 & \textbf{0.0019} & 0.8095 & 0.9977 & \textbf{0.8288} & Ours \\
\\
\multirow{5}{*}{Mixed} & 0.3015 & 0.0225 & 0.1294 & 0.9878 & 0.1618 & PG-LSR \\
                       & 0.5388 & 0.0918 & 0.7114 & 0.9963 & 0.5783 & ID \\
                       & 0.6424 & 0.2013 & 0.0815 & 0.9890 & 0.1409 & WT \\
                       & 0.3821 & 0.0069 & \textbf{0.8526} & 0.9832 & 0.5027 & PN-RPCA \\
                       & 0.7014 & 0.0586 & 0.2194 & 0.9897 & 0.3277 & SOMC \\
                       & \textbf{0.8514} & \textbf{0.0025} & 0.8489 & \textbf{0.9974} & \textbf{0.8496} & Ours \\
\bottomrule
\end{tabular}
\end{table}

\begin{figure}[!t]
	\centering
	\includegraphics[width=4.5in]{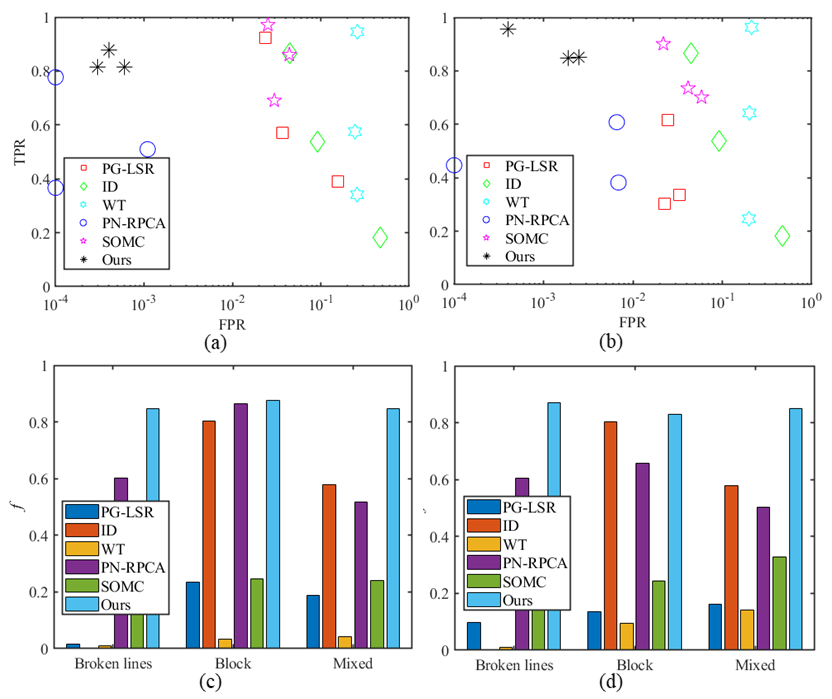}
	\caption{TPR-FPR scatter plot of (a) square metallic mesh and (b) circular metallic mesh, f-value histogram of (c) square metallic mesh and (d) circular metallic mesh.}
	\label{fig_8}
\end{figure}

\begin{figure}[!t]
	\centering
	\includegraphics[width=5.5in]{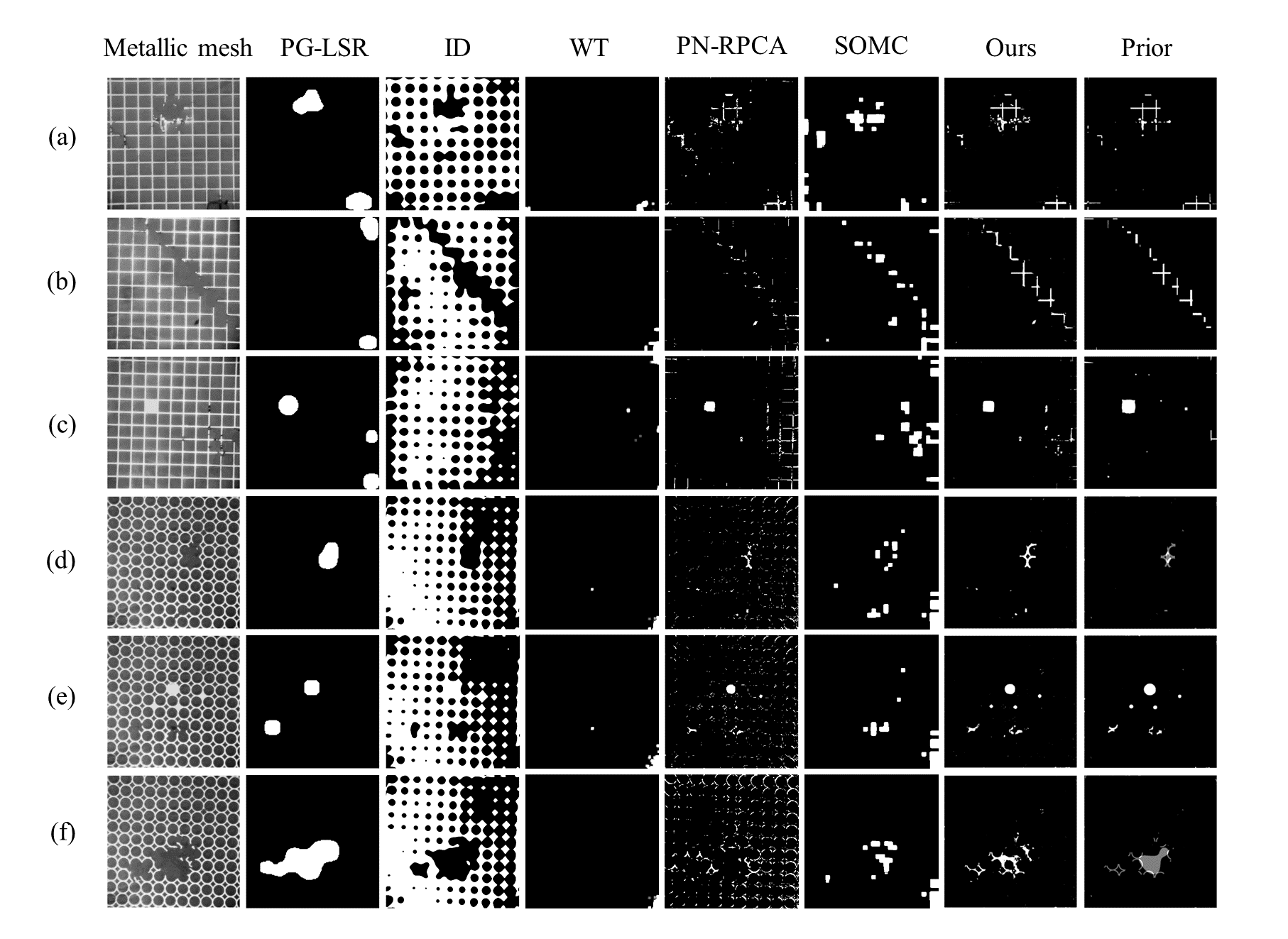}
	\caption{The acquired image of square metallic mesh and circular metallic mesh with the comparison of multiple defect detection results.}
	\label{fig_9}
\end{figure}

\begin{figure}[!t]
	\centering
	\includegraphics[width=3.5in]{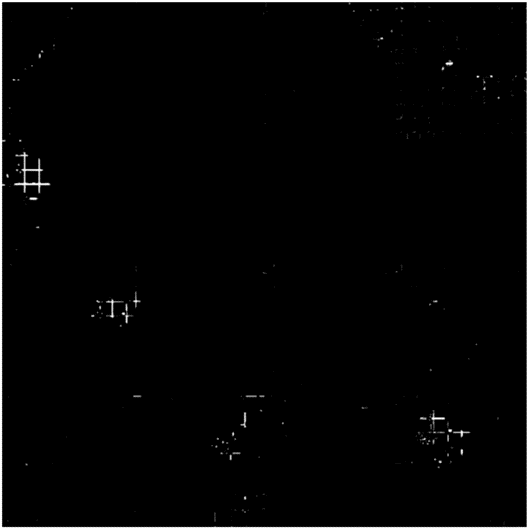}
	\caption{Detection results of metallic mesh defects beyond the field of view range.}
	\label{fig_10}
\end{figure}
The TPR-FPR scatter plot and $f$-value histogram are shown in Fig. 8. In the scatter plot, the upper left corner is most desirable, indicating a higher TPR and a lower FPR, signifying superior defect detection performance by the algorithm. Regarding the $f$-value histogram, higher values correlate with better performance.

The scatter distribution for PG-LSR, WT, ID and SOMC is notably scattered and far from the vertical axis, markedly underperforming compared to DW-RPCA. 
The scatter distributions of PN-RPCA, while closer to the vertical axis, suggest higher FPRs and worse TPRs compared to DW-RPCA. 
In the $f$-value histogram for the square and circular metallic mesh, PG-LSR, ID, WT, SOMC performs poorly, PN-RPCA achieves good results, and DW-RPCA achieves optimal $f$-value for all three types of defective image sets.

\subsection{Experimental results of LD-RSM}

Square metallic mesh and circular metallic mesh images have been acquired by the proposed LD-RSM system to validate the performance of DW-RPCA.
Note that ground truth are generally not available due to the large number and complexity of the actual collection scenarios.
The acquired image of square metallic mesh and circular metallic mesh with the comparison of multiple defect detection results are shown in Fig. 9.
Columns 2-5 show the PG-LSR, ID, WT, PN-RPCA, SOMC, and DW-RPCA assays used for comparison and prior information as in columns 6 and 7, respectively.
It can be found that PG-LSR has a certain detection effect, however, the method is more serious misdetection phenomenon, and is ineffective for the detection of defects with a small area.
ID detects defects with a large-area misdetection, and this paper analyzes that the problem is caused by uneven light intensity and image noise. 
WT also fails to detect.
PN-RPCA can basically realize the detection of defects, but the performance in real images is not as good as before, due to the insufficient anti-interference ability of the model.
SOMC can detect defects, but there are some false positives.
DW-RPCA is robust to uneven illumination, noise and other problems, and can realize fast and accurate detection for three types of defects of different sizes and shapes, effectively improving the problems of misdetection and leakage.

DW-RPCA is utilized to detect defects in the acquired image of a specific region (2 mm $\times$ 2 mm) by LD-RSM, and the defect detection results are shown in Fig. 10. Defect detection results are accurate, indicating that the metallic mesh defect detection system constructed in this paper, by scanning the image based on the UR5 robotic arm and LD-RSM, and then using DW-RPCA for detection, can realize the ultra-field-of-view range of the metallic mesh specific areas of fast and accurate detection, with the defects of the metallic mesh fine inspection capabilities.

\section{Conclusion}
In this paper, a reflective portable imaging system (LD-RSM) is designed to realize the imaging in the in-situ state of the metallic mesh. 
The optical path structure coupled with an external optical lens set and a smartphone camera module is adopted to increase the working distance (22.23mm) of the imaging system.
The introduction of an aperture diaphragm and other stray light filtering designs effectively reduces the impact of stray light on the imaging quality of the imaging system. 
%The external optical lens group is encapsulated using 3D printing, and the dimensions of the optical assembly are 118 mm × 122 mm × 50 mm. 
Experimental results show that the resolution of the reflective portable imaging system is 4.92 $\mu$m, and the diameter of the field of view is 800 $\mu$m. 
A low-rank decomposition defect detection algorithm based on the fusion of spectral filtering and the Hoff Transform (DW-RPCA) is also designed to acheive the detection of defects in the square and circular metallic mesh. 
DW-RPCA has the best detection results compared to other methods, achieving f-numbers 0.856 in square metallic mesh and 0.848 in circular metallic mesh.
Finally, the performance of LD-RSM and DW-RPCA was demonstrated in combination with a robotic arm to realize the ultra-field-of-view inspection of large-size micro- and nanostructures. It provides a low-cost and efficient solution for industrial in-situ inspection.

\appendix
\section{Declaration of Competing Interest}
Authors declare that they have no conflict of interest. 

\section{CRediT authorship contribution statement}
\textbf{Zhengang Lu: }Supervision, Conceptualization, Methodology, Validation, Writing – original draft, Writing – review $\&$ editing. 
\textbf{Hongsheng Qin: }Investigation, Visualization, Formal analysis, Data curation, Investigation. 
\textbf{Jing Li: }Software, Data curation, Resources, Validation. 
\textbf{Ming Sun: }Writing – original draft, Writing – review $\&$ editing.
\textbf{Jiubin Tan: }Supervision, Writing – review $\&$ editing.

\section{Acknowledgments}
This work was supported by the National Natural Science Foundation of China under Grant No.61975046. 

\bibliographystyle{cas-model2-names}

\bibliography{qhs2/main-refs}

\end{document}